\documentclass[11pt]{article}

\usepackage[margin=1in]{geometry}
\usepackage{amsmath,amssymb,amsthm}
\usepackage{graphicx}
\usepackage{hyperref}
\usepackage{natbib}
\usepackage{authblk}

\title{A Lagrangian View of Flow Matching}
\author[1]{Peyman Milanfar}
\affil[1]{Google}
\date{September 4, 2026}

\begin{document}
\maketitle

\begin{abstract}
Modern explicit-time generative models, such as Flow Matching \citep{lipman2023flow} and Rectified
Flow \citep{liu2023flow}, are typically derived top-down via Optimal Transport and the continuity
equation. This standard Eulerian approach focuses on the macroscopic transport of probability mass.
In this paper, we present an alternative, bottom-up mechanical derivation grounded in a Lagrangian
(particle-centric) perspective. By analyzing the local Taylor expansion of a continuous denoiser, we
motivate a strict invariance condition required for optimal, single-step generation: the conservation
of target identity. Enforcing this condition yields a governing quasi-linear advection Partial
Differential Equation (PDE). We demonstrate that solving this PDE via the Method of Characteristics
analytically yields the straight-line trajectories of Flow Matching. This geometric perspective
isolates the Jacobian of the denoiser as the primary source of trajectory curvature, providing a
direct mathematical explanation for why straight-line flows enable massive step sizes, and why
empirical models require distillation to flatten intersecting characteristics.
\end{abstract}

\section{The Predictor-Corrector and the Moving Target}

Consider a generative process mapping noise to data over a continuous time parameter $t \in [0,1]$,
where $t=0$ defines the clean data manifold. Let $x(t)$ be the state of the trajectory. We rely on a
differentiable denoising function $f(x,t)$ whose objective is to predict the final clean data state
$x_0$ from the current state $x(t)$.

As the state moves infinitesimally along the trajectory by $dx$ over time $dt$, the denoiser's
prediction of the target changes according to its total continuous differential:
\begin{equation}
df = J_f(x,t)\,dx + \frac{\partial f}{\partial t}(x,t)\,dt
\label{eq:1}
\end{equation}
where $J_f$ is the spatial Jacobian matrix and $\partial f/\partial t$ represents the temporal drift.

This continuous drift presents a fundamental challenge for the discrete ODE solvers used in standard
generative models (such as Gaussian diffusion). To understand this algorithmic penalty, consider what
happens when a discrete solver attempts to navigate this changing field across a single, finite time
step. Suppose the solver is currently at state $x$ at time $t$:

\begin{enumerate}
\item \textbf{The Predictor:} The solver applies the denoiser to get its first prediction,
$f(x,t)$. It calculates the residual error between its current state and this target,
$(x - f(x,t))$, and uses this to dictate the discrete step forward, $\Delta x$.
\item \textbf{The Corrector:} The solver arrives at the new state, $x+\Delta x$, at the advanced time
$t - \Delta t$. It must now apply the denoiser a second time to re-evaluate the target.
\end{enumerate}

We can observe the denoiser's behavior during this second application by taking a first-order Taylor
expansion at the new state, which acts as the discrete, macroscopic version of the total differential:
\begin{equation}
f(x+\Delta x, t-\Delta t) \approx f(x,t) + J_f(x,t)\Delta x - \frac{\partial f}{\partial t}(x,t)\Delta t
\label{eq:2}
\end{equation}

This expansion reveals a fundamental bottleneck. Because the noise schedule curves the underlying
manifold, the destination is constantly shifting. The Jacobian $J_f$ corrects the estimate by
reacting to the newly revealed local geometry at $x+\Delta x$, while the temporal derivative accounts
for the receding timeline. Consequently, in standard diffusion, the denoiser's prediction of the
clean image changes at every single step. Because the target is continuously moving, solvers are
forced to take extremely small, conservative step sizes ($\Delta t$) to avoid overshooting the
shifting destination.

\section{The Invariance Principle: From Continuity to Advection}

If the moving target is the bottleneck for fast generation, the theoretical ideal is a trajectory
where the target never moves. Suppose we could design a ``perfect'' generative trajectory $x(t)$. On
this path, the denoiser would correctly identify the exact clean image $x_0$ at the very first noise
level, and as we step forward in time, that prediction would never waver. Mathematically, this
requires the denoiser's output to be strictly invariant along the trajectory:
\begin{equation}
f(x(t_1), t_1) = f(x(t_2), t_2) = x_0
\label{eq:3}
\end{equation}

For this to hold continuously, the total rate of change of the denoiser's output along the path
$x(t)$ must be exactly zero:
\begin{equation}
\frac{d}{dt} f(x(t), t) = 0
\label{eq:4}
\end{equation}

This consistency constraint has been used elsewhere \citep{geng2025consistency} to construct a
finite-difference training loss. Here however, we use it to introduce a complete shift in
perspective. Standard Flow Matching and stochastic interpolant literature
\citep{lipman2023flow, albergo2023building} derives the velocity field from an Eulerian
(field-centric) perspective, relying on the Continuity Equation
($\partial p_t/\partial t + \nabla\cdot(p_t v_t) = 0$) to dictate the conservation of probability mass
across space.

In contrast, Equation~\eqref{eq:3} represents a Lagrangian (particle-centric) perspective. Rather than
standing at a fixed coordinate watching probability mass flow past, we are conceptually sitting inside
a specific particle $x(t)$ riding the flow. Geometrically, the ideal denoiser acts as the inverse flow
map ($f(x,t) = \phi_t^{-1}(x) = x_0$). Thus, our invariance principle demands that the target identity
(the origin $x_0$) of the particle is perfectly conserved as it flows.

Applying the multivariate chain rule, we expand this total derivative into its spatial and temporal
components:
\begin{equation}
\frac{df}{dt} = \frac{\partial f}{\partial t}(x,t) + J_f(x,t)\,\frac{dx}{dt} = 0
\label{eq:5}
\end{equation}

Let $dx/dt$ be the velocity vector field $v(x,t)$ driving the generative state. Substituting this, we
arrive at the governing equation for an invariant generative field:
\begin{equation}
\frac{\partial f}{\partial t}(x,t) + J_f(x,t)\,v(x,t) = 0
\label{eq:6}
\end{equation}

Equation~\eqref{eq:6} is a system of first-order, quasi-linear advection PDEs. It dictates that any
temporal drift in the target manifold ($\partial f/\partial t$) must be perfectly neutralized by
advecting the state through the local geometry ($J_f$) of the denoiser. While the Eulerian continuity
equation describes the same underlying dynamics at macro scale, this Lagrangian PDE isolates the local
spatial Jacobian, making it well suited to analyze the mechanical difficulties of discrete ODE solvers.

\section{Solving the PDE: Deriving Flow Matching}

Rather than relying on Optimal Transport, the Conservation Equation, or Fokker--Planck marginals, we
can construct a generative model by directly designing a velocity field $v(x,t)$ that solves this PDE.
To do so, we employ the Method of Characteristics.

The governing PDE \eqref{eq:6} constrains the pair $(f,v)$ jointly, but does not by itself pick out a
unique velocity field: for an arbitrary denoiser $f$, infinitely many fields $v$ satisfy
$\partial_t f + J_f v = 0$ at a given point. To make progress, we therefore \emph{posit} a candidate
velocity field, motivated by Tweedie's formula \citep{efron2011tweedie} and the residual-driven flow
of \citet{milanfar2024denoising}, rather than deriving it from the PDE alone. Recall Tweedie's
formula-based residual flow,
\begin{equation}
\frac{dx_t}{dt} = \frac{1}{2}\frac{d\alpha_t}{dt}\,\frac{(x_t - \mathbb{E}[x_0\mid x_t])}{\alpha_t},
\label{eq:7}
\end{equation}
where the velocity is driven entirely by the residual error between the current state and the optimal
MMSE denoiser. Generalizing the ideal denoiser $\mathbb{E}[x_0\mid x_t]$ to our continuous denoiser
$f(x,t)$, and absorbing the noise-schedule coefficients into a single arbitrary scalar scheduling
function $c(t)$, we postulate the residual-based velocity field
\begin{equation}
v(x,t) = c(t)\big(x - f(x,t)\big).
\label{eq:8}
\end{equation}

The role of the Method of Characteristics here is \emph{not} to derive Equation~\eqref{eq:8}, but to
test whether this specific ansatz can be made self-consistent with the invariance principle of
Section~2, and --- if so --- to solve for the resulting trajectories.

We evaluate the denoiser along a curve $X(t)$ whose velocity is \emph{defined} to match our ansatz,
\begin{equation}
\frac{dX(t)}{dt} = c(t)\big(X(t) - f(X(t), t)\big).
\end{equation}
By the chain rule this gives, identically and for \emph{any} choice of $c(t)$,
\begin{equation}
\frac{d}{dt} f(X(t), t) = \frac{\partial f}{\partial t}(X(t), t) + J_f(X(t), t)\, c(t)\big(X(t) -
f(X(t), t)\big).
\label{eq:9}
\end{equation}
Note that Equation~\eqref{eq:9} is not yet a statement about invariance --- it is simply the chain
rule applied along $X(t)$. The content of our construction lies entirely in \emph{imposing} the
invariance condition on this specific curve, i.e.\ requiring
\begin{equation}
\frac{d}{dt} f(X(t), t) = 0
\label{eq:9prime}
\end{equation}
along $X(t)$. This is a choice, not a consequence of Equation~\eqref{eq:8} alone --- it is the same
Section~2 invariance principle, now applied to the particular trajectory generated by our ansatz.
Equation~\eqref{eq:9prime} tells us that \emph{if} such a self-consistent trajectory exists,
$f(X(t),t)$ must be constant along it. The boundary condition $f(x,0) = x_0$ then fixes that constant:
\begin{equation}
f(X(t), t) = x_0 \quad \text{for all } t.
\label{eq:10}
\end{equation}

This is now a genuine consequence we can exploit. Substituting the constant target \eqref{eq:10} back
into the ansatz \eqref{eq:8} turns the coupled PDE system into a single, closed, separable ODE in
$X(t)$ alone:
\begin{equation}
\frac{dX(t)}{dt} = c(t)\big(X(t) - x_0\big).
\label{eq:11}
\end{equation}
This is now a separable, linear ODE. Integrating both sides with respect to time yields:
\begin{align}
\int \frac{1}{X(t) - x_0}\, dX &= \int c(t)\, dt
\label{eq:12}\\
\ln|X(t) - x_0| &= \ln \sigma(t) + \text{const}
\label{eq:13}\\
X(t) &= x_0 + \sigma(t)\, z_1
\label{eq:14}
\end{align}
where $\sigma(t) = \exp\!\big(\int c(t)\, dt\big)$ serves as the noise schedule, and $z_1$ is the
constant of integration representing the initial noise state at $t=1$.

Equation~\eqref{eq:14} is the exact formulation of Flow Matching and Rectified Flow
\citep{lipman2023flow, liu2023flow}. By requiring the velocity field to be reconciled with the
generative advection PDE, the only self-consistent characteristic curves under this residual ansatz
are straight lines radiating from the data $x_0$ to the noise $z_1$.

\paragraph{What this derivation shows, precisely.} We have not proven that the residual ansatz
\eqref{eq:8} is the \emph{unique} velocity field solving the invariance PDE --- other choices of $v$
could in principle also be made self-consistent. What we have shown is narrower, and we believe still
instructive: (i) the residual ansatz \emph{can} be reconciled with the invariance principle,
(ii) doing so forces $f$ to be constant along the resulting characteristics, and (iii) that single
requirement, combined with the boundary condition at $t=0$, is sufficient to pin down the trajectories
completely --- and they are exactly the straight lines of Flow Matching. The Method of Characteristics
does real work in step (iii); the ansatz in Equation~\eqref{eq:8} does the work of narrowing which PDE
solution we look for in the first place.

\section{The Jacobian Penalty and Trajectory Curvature}

This PDE framework cleanly isolates why standard diffusion models require thousands of solver steps,
while Flow Matching requires very few. We can rearrange the advection PDE to solve for the temporal
drift of the target:
\begin{equation}
\frac{\partial f}{\partial t}(x,t) = -J_f(x,t)\, v(x,t)
\label{eq:15}
\end{equation}

In standard diffusion models, the noise schedules non-linearly warp the intermediate marginal
distributions. Because the probability mass diffuses non-linearly over time, the optimal target
prediction inherently shifts at every timestep. Consequently, standard diffusion trajectories do not
lie on the straight-line characteristics of our PDE.

When a trajectory deviates from the characteristic curve, the temporal drift $\partial f/\partial t$
becomes heavily non-zero. Equation~\eqref{eq:15} shows that tracking this shifting target requires
continuously multiplying the velocity by the Jacobian matrix $J_f$.

To understand the physical mechanics of this drift, we can view the denoiser through a statistical
lens. For an optimal MMSE denoiser (or an ``ideal'' denoiser with a symmetric Jacobian), the spatial
Jacobian is directly proportional to the posterior covariance matrix ($\Sigma_{\text{post}}$) of the
signal given the noisy observation \citep{manor2024posterior}: $J_f(x,t) \propto \Sigma_{\text{post}}$.

Because any valid covariance matrix is Positive Semi-Definite (PSD), this imposes a strict geometric
constraint on the target drift. If we measure how the target shifts along the direction of travel by
taking the dot product of Equation~\eqref{eq:15} with the velocity $v(x,t)$, we find:
\begin{equation}
v(x,t)^\top \frac{\partial f}{\partial t} = -v(x,t)^\top J_f(x,t)\, v(x,t) \le 0
\label{eq:16}
\end{equation}

This provides a clear demonstration of solver instability: the temporal drift of the target is
guaranteed to be negatively correlated with the velocity. The target will always retreat or resist the
direction of travel, forcing standard curved ODE solvers to take infinitesimal steps to avoid
overshooting. Furthermore, the target drifts exactly along the principal axes of the model's internal
uncertainty ($\Sigma_{\text{post}}$).

Geometrically, as $t \to 0$, the target data manifold becomes infinitely sharp, causing the
eigenvalues of the spatial Jacobian $J_f$ (i.e., posterior covariance) to diverge. If the trajectory is
curved ($\partial f/\partial t \ne 0$), discrete ODE solvers are subjected to this exploding Jacobian,
requiring infinitesimal step sizes to maintain stability.

Conversely, because Flow Matching analytically solves the PDE along straight characteristics, the
target is static ($\partial f/\partial t = 0$). The Jacobian penalty is mathematically neutralized,
allowing the solver to confidently traverse the space in massive step sizes without being thrown off
course by the local geometry.

\section{The Engineering Reality: Intersecting Characteristics}

While the analytical solution to the PDE yields straight lines, empirical training of neural networks
introduces a fundamental violation of the PDE's assumptions.

During the training of a Flow Matching model, straight characteristic lines are drawn independently
between random noise vectors $z_1$ and data samples $x_0$. In high-dimensional space, the path to one
data point may physically intersect with the path to another.

At an intersection point $x_{\text{int}}$, the PDE demands two different target values for
$f(x_{\text{int}}, t)$, which is mathematically impossible for a deterministic function. To minimize
loss, a neural network learns to output the expected (average) velocity at these intersections
\citep{liu2023flow}. Statistically, this intersection represents a point of extreme ambiguity where the
model's internal belief system is split. Consequently, the posterior covariance $\Sigma_{\text{post}}$
explodes at this coordinate, exhibiting massive eigenvalues along the principal axes spanning the
conflicting targets.

This marginalization warps the vector field, bending the learned trajectories off the perfect
characteristic curves. Consequently, $\partial f/\partial t$ is no longer zero, the Jacobian penalty
returns, and baseline Flow Matching models still require intermediate solver steps (e.g., 20 to 50
steps) to navigate the resulting curvature.

To achieve true single-step generation, the characteristics must be uncrossed. This is achieved
through Reflow (distillation) \citep{liu2023flow}. A baseline model simulates valid, non-intersecting
trajectories from $t=1$ to $t=0$. A new model is then trained to draw straight lines between these
pre-matched pairs. Viewed through our statistical lens, Reflow is fundamentally an uncertainty
elimination protocol. By untangling the paths, the mapping from any point on the trajectory to the
final data point becomes deterministic. Without conflicting targets, the model's posterior uncertainty
drops to zero, causing the covariance matrix to vanish ($\Sigma_{\text{post}} \to 0$). Because the
covariance vanishes, the spatial Jacobian becomes a zero matrix ($J_f \to 0$), completely starving the
mechanism that generates target drift. This process effectively constructs a true analytical solution
to the advection PDE where characteristics never cross, driving $\partial f/\partial t \to 0$ globally
and allowing true single-step evaluation.

\subsection{A Minimal Working Example: The Two-Mode Crossing}
\label{sec:toy-example}

To make the abstract claims of this section concrete, consider the simplest possible setting where
characteristics cross: a 1D data distribution with two equally likely modes, $x_0 \in \{-1, +1\}$,
coupled to noise via the standard linear interpolation $x_t = (1-t)x_0 + t z_1$, $z_1 \sim
\mathcal{N}(0,1)$. Individually, the two characteristic families $X(t) = (1-t)(\pm 1) + t z_1$ are
straight lines by construction. But because both modes share the same noise distribution, lines
originating from $x_0 = -1$ and $x_0 = +1$ densely overlap near $x=0$ for $t$ close to $1$, and the
Bayes-optimal (i.e., population-optimal) denoiser --- exactly what a well-trained network converges to
--- must average over both explanations there. For this symmetric two-point Gaussian mixture, the
posterior mean has closed form:
\begin{equation}
f(x,t) = \tanh\!\left(\frac{(1-t)\,x}{t^2}\right), \qquad v(x,t) = \frac{x - f(x,t)}{t}.
\label{eq:17}
\end{equation}

\begin{figure}[htbp]
\centering
\includegraphics[width=0.62\textwidth]{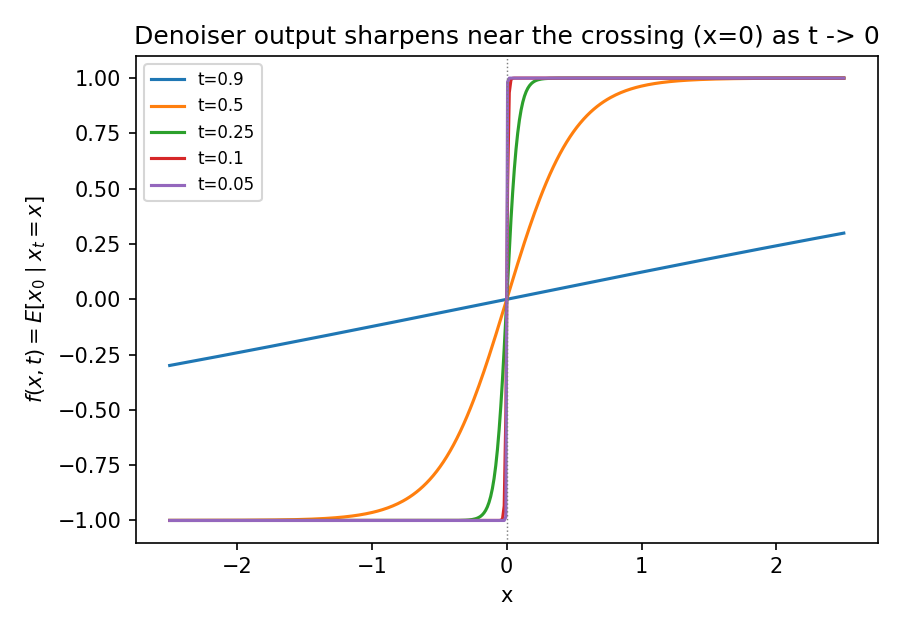}
\caption{The denoiser output $f(x,t)$ across several noise levels $t$. At $t=0.9$ it is nearly linear
(heavy averaging, low confidence); by $t=0.1$ it has collapsed into a step function at $x=0$,
confirming that the Jacobian $J_f = \partial f/\partial x$ diverges as $t \to 0$ precisely at the
crossing coordinate, as predicted in Section~4.}
\label{fig:denoiser-sharpening}
\end{figure}

\begin{figure}[htbp]
\centering
\includegraphics[width=0.62\textwidth]{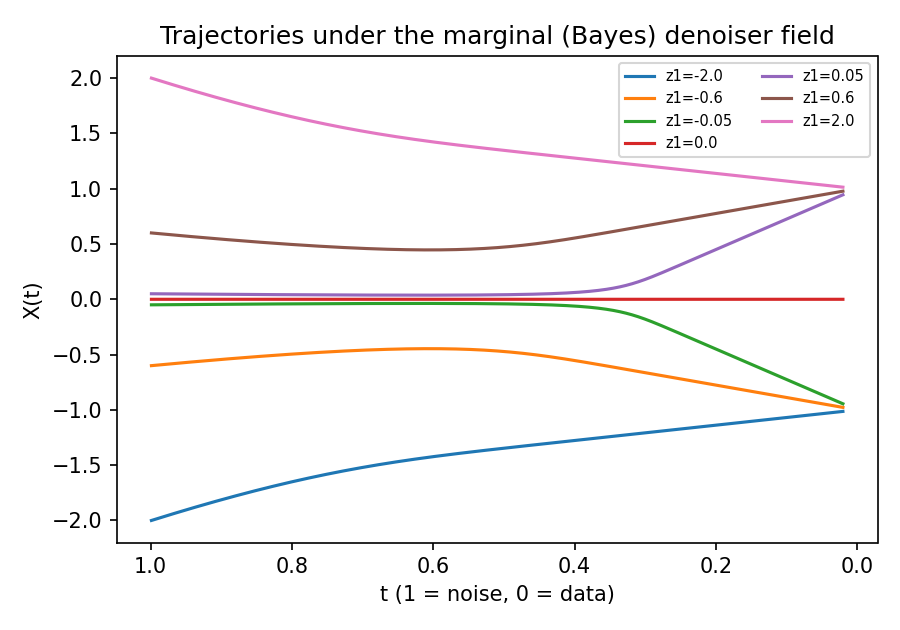}
\caption{Trajectories $X(t)$ obtained by integrating $\dot X = v(X,t)$ backward from $t=1$ (noise) to
$t \approx 0$ (data), for several initial draws $z_1$. Distant starts ($|z_1|$ large) are nearly
straight; starts near the crossing ($z_1 \approx 0$) show pronounced curvature, and the degenerate case
$z_1 = 0$ never resolves at all.}
\label{fig:trajectories}
\end{figure}

Figure~\ref{fig:denoiser-sharpening} plots $f(x,t)$ across $t$, visually confirming the exploding-Jacobian
argument of Section~4. Figure~\ref{fig:trajectories} integrates $\dot X = v(X,t)$ backward from $t=1$
to $t \approx 0$ for several draws of $z_1$. Trajectories starting far from the crossing
($z_1 = \pm 2.0$) are nearly straight, matching the single-mode analytical solution of Section~3. But
trajectories seeded close to the decision boundary ($z_1 = \pm 0.05$) are visibly curved S-shapes: they
linger near $x \approx 0$ before being pulled decisively toward one mode only in the final stretch ---
exactly the ``characteristics bending off course near an intersection'' mechanism described
qualitatively above.

\begin{figure}[htbp]
\centering
\includegraphics[width=0.62\textwidth]{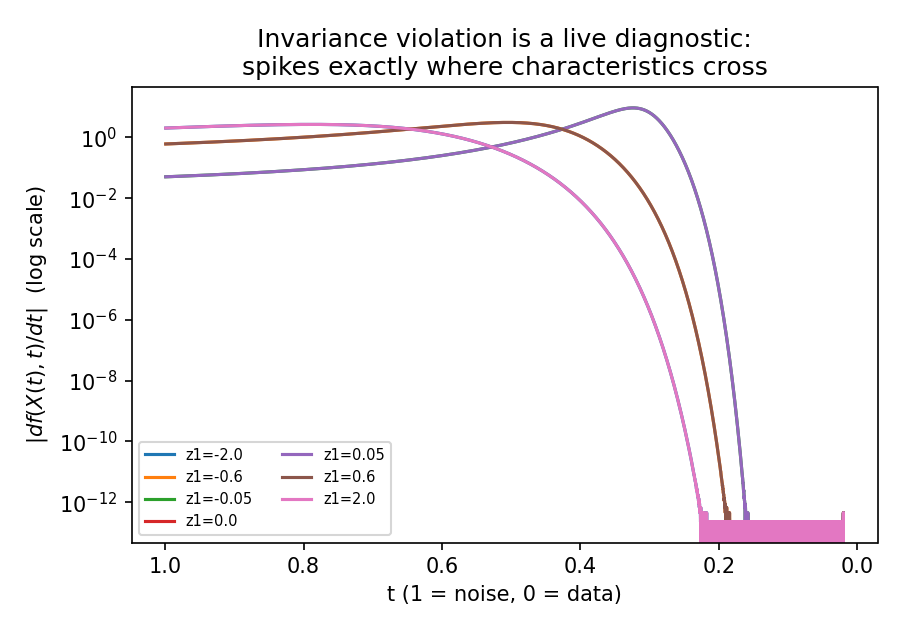}
\caption{The invariance-violation diagnostic $|df(X(t),t)/dt|$ (log scale) along each trajectory of
Figure~\ref{fig:trajectories}. Trajectories seeded near the crossing sustain large drift far longer
than distant trajectories, directly localizing where the advection PDE is being violated.}
\label{fig:drift-diagnostic}
\end{figure}

Figure~\ref{fig:drift-diagnostic} turns this into the Section~\ref{sec:diagnostic} diagnostic directly:
it plots $|df(X(t),t)/dt|$ along each trajectory. For distant starts, this invariance-violation measure
decays to numerical zero well before $t=0$ (the target has been resolved and stops moving --- the PDE
is satisfied). For near-crossing starts, it stays orders of magnitude larger for longer, peaking around
$t \approx 0.2$--$0.5$ before finally collapsing. Numerically, the near-crossing trajectories
($z_1 = \pm 0.05$) exhibit a peak drift roughly $3.5\times$ larger than the far trajectories
($z_1 = \pm 2.0$): $\max|df/dt| \approx 9.23$ vs.\ $\approx 2.65$.

This confirms the practical proposal of Section~\ref{sec:diagnostic}: the instantaneous quantity
$|df(X(t),t)/dt|$, computed directly from consecutive denoiser evaluations during sampling --- with no
access to ground truth needed --- is a real-time, per-sample signal for exactly where a trajectory is
fighting unresolved crossings, and hence exactly where a solver should slow down (or where
Reflow-style distillation would most benefit that region of the data manifold). Note also the
degenerate case $z_1 = 0$ in Figures~\ref{fig:trajectories}--\ref{fig:drift-diagnostic}: starting
exactly on the decision boundary, the trajectory never resolves at all --- $f$ stays frozen at the
unstable fixed point $f=0$, a toy analogue of the mode-averaging/blur failure mode seen in low-step
generative samples.

This example is deliberately the simplest possible case (two modes, one dimension, closed-form
posterior). Real data manifolds have continuously many ``modes,'' so in practice the diagnostic would
be evaluated as a distribution over many samples rather than a handful of trajectories, but the
underlying mechanism scales directly.

\section{Discussion and Practical Implications}

While the standard Eulerian perspective (e.g., Optimal Transport, Fokker--Planck) provides rigorous
guarantees that the overall distribution of generated images matches the training data, it remains a
macroscopic view. It abstracts away the discrete mechanics of the actual algorithms running on
hardware. The Lagrangian (particle-centric) perspective presented in this paper offers several
concrete, practical advantages for designing and debugging generative models:

\subsection{A Tangible Metric for Inference Monitoring}
\label{sec:diagnostic}

The Eulerian view deals with vector fields ($v_t$) and probability masses ($p_t$), which are abstract
and cannot be directly measured during generation. Neural networks do not output probability masses;
they output a tensor of pixel values (the denoiser prediction, $f(x,t)$). Our derivation is built
entirely around $f(x,t)$ --- the exact quantity the neural network computes. This implies that the core
mathematical invariant ($\frac{d}{dt}f(x,t) = 0$) can be directly monitored during inference. If we
monitor the neural network's output across solver steps and observe wild fluctuations, the advection
PDE is being violated, immediately indicating that the trajectory is curved and smaller step sizes (or
further distillation) are required. The toy example of Section~\ref{sec:toy-example} demonstrates this
diagnostic concretely.

\subsection{Reframing Distillation as Intersecting Characteristics}

In standard literature, ``Reflow'' or distillation is often described abstractly as straightening the
transport couplings between distributions. The Lagrangian view reframes this as intersecting paths. If
a neural network is forced to evaluate a region where two particle paths cross, it must average the
conflicting velocities. This averaging bends the flow, shifts the target, and immediately reactivates
the Jacobian penalty. This provides a highly intuitive diagnostic lens: if a 1-step generator produces
blurry results, it is mechanically because its characteristic curves are still crossing in
high-dimensional space, requiring further Reflow to untangle.

\subsection{Relation to Flow Maps}
\label{sec:flow-maps}

A recent and far more general line of work, Flow Map Matching (FMM) \citep{boffi2024flowmap}, learns
the two-time flow map $X_{s,t}$ of a probability flow ODE: a function satisfying the jump condition
$X_{s,t}(x_s) = x_t$, which generalizes one-step consistency models to a bidirectional map supporting
arbitrary post-training discretization \citep{boffi2025howto}. 

In the notation of \citet{boffi2024flowmap}, our
continuous denoiser is exactly $f(x,t) = X_{t,0}(x)$: the flow map from the current time back to the
clean-data endpoint, with the destination time fixed at $0$. The general flow-map consistency law is
the semigroup property $X_{s,t} \circ X_{r,s} = X_{r,t}$, whose infinitesimal (generator) form is
$\partial_t X_{s,t}(x) = b_t(X_{s,t}(x))$. Fixing the target time at $s=0$ and renaming
$X_{t,0} \to f(\cdot,t)$ recovers precisely our governing PDE, Equation~\eqref{eq:6}. In this sense,
the invariance principle of Section~2 and the Method-of-Characteristics derivation of Section~3 amount
to an elementary, from-scratch re-derivation (via Taylor expansion rather than semigroup theory) of the flow-map consistency law restricted to the single-target-time slice $X_{\cdot,0}$. This places
our simple construction thusly: consistency functions (a single fixed
source and target time) sit inside our time-varying denoiser $f(x,t)$ (one free time, fixed target),
which in turn sits inside the full two-time flow map $X_{s,t}$ (both times free).

It's worth clarifying a subtle but important difference in terminology: Boffi et al. \citet{boffi2024flowmap} introduces a Lagrangian \textit{objective} for distilling a flow map from a pre-trained velocity field, complementary to a related Eulerian loss that they prove is the continuous-time limit of consistency distillation, and
show that both control the Wasserstein distance between teacher and student. This is the same
Eulerian/Lagrangian vocabulary organizing the present paper, but the two dichotomies are not identical
and should not be conflated. Their ``Lagrangian'' loss integrates a self-distillation residual along
solver trajectories, versus an ``Eulerian'' loss enforcing instantaneous self-consistency at a point, whereas our Eulerian/Lagrangian split instead contrasts probability-mass transport with single-particle target-tracking (Equation~\eqref{eq:4}). The two usages rhyme, but they formalize different distinctions within the same underlying dynamical system.

Finally, while FMM's account of why direct
(non-distillation) flow-map training is difficult relies on Lipschitz and Wasserstein control bounds, Section~4 of the present paper offers a complementary, more
physically interpretable explanation of the same phenomenon: the identification $J_f \propto
\Sigma_{\text{post}}$ together with the PSD sign constraint of Equation~\eqref{eq:16} says concretely
\emph{where} and \emph{why} the map's Jacobian misbehaves.

It is worth noting that the inference-time diagnostic of Section~\ref{sec:diagnostic},
$|df(X(t),t)/dt|$, generalizes directly to FMM's two-time setting. For a trained flow map $X_{s,t}$, the
analogous online consistency check is the jump-condition residual
\begin{equation}
r(s,t,x) = \big\| \partial_t X_{s,t}(x) - b_t(X_{s,t}(x)) \big\|,
\label{eq:18}
\end{equation}
which vanishes for a perfect map and spikes wherever $X_{s,t}$ is forced to average over crossing
characteristics.

\bibliographystyle{plainnat}

\end{document}